# R-C-P Method: An Autonomous Volume Calculation Method Using Image Processing and Machine Vision

MA Muktadir , Sydney Parker, and Sun Yi

**Abstract:** Machine vision and image processing are often used with sensors for situational awareness in autonomous systems, including industrial robots and self-driving cars. In the domain of autonomous systems , 3D depth sensors, such as LiDAR and Radar, prove to be great tools, transforming the way these systems perceive and interact with their surroundings. However, incorporating 3D sensors into the system results in increased power consumption, larger volumes of data, additional cabling, physical space, and heightened temperature and mechanical requirements for operation. Therefore, these sensors might not be suitable for certain operational settings, such as a space environment. Conversely, a 2D camera presents a cost-effective and straight forward solution in terms of both installation and operation. This study was motivated by a desire to obtain real-time volumetric data and study information with multiple 2D cameras instead of a singular depth camera. In this venture, two cameras were used to measure the dimensions of a rectangular object in real time. During this experiment, the row-column-pixel (R-C-P) method was conducted using a 2D camera and three rectangular surfaces at the same distance. The R-C-P method is developed by using image processing and edge detection. In addition to the surface areas, the R-C-P method also detects discontinuous edge volume. The relationship between the actual dimension and the dimension output by the R-C-P method is established as the distance gradually increases. Altogether, the obtained results best illustrated the R-C-P method, providing equations to calculate the rectangular object's surface area dimensions. By inputting the measured distance between the object and the camera, the vision system provides the dimensions of actual objects.

## 1. Introduction

Two and three-dimensional machine vision systems can provide three-dimensional information about an object. However, specific scenarios exist where the feasibility of 3D estimation becomes elusive or presents inherent challenges. Highlighted are several compelling reasons advocating the favorability of 2D techniques. In light of the potential expense associated with high-quality 3D cameras, opting for 2D cameras or alternative solutions becomes a more pragmatic choice in budget-constrained circumstances. In addition to be cost-friendly, the practicality of 2D cameras becomes evident in compact or confined spaces, where the installation of 3D scanning equipment is challenging, and maneuvering for a comprehensive 3D view is impractical. Additionally, translucent, or highly reflective surfaces pose challenges for specific 3D scanning systems. Under such circumstances, it becomes difficult to obtain precise depth information, therefore 2D techniques can be more appropriate. Real-time processing or instantaneous data acquisition is necessary for some systems. Some 3D scanning techniques may lack the requisite speed leading to the use of a significant amount of power. In scenarios where time or processing systems are limited, simpler 2D techniques may be favored. For instance, basic volume estimations may be sufficient. Altogether, specific applications may not necessitate precise volumetric measurements and can suffice with 2D approximations.

Numerous touchless monitoring systems require volumetric data, especially in applications such as object surveillance and defect detection. For example, a faster R-CNN-based method was developed for detecting concrete spalling damage using depth cameras or 3D scanners [1]. Another study proposes a deep learning-based approach for detecting and reconstructing buildings from a single aerial image. The method employs an optimized multi-scale convolutional–deconvolutional network to acquire the necessary information for reconstructing the 3D shapes of buildings, including height data and linear elements of individual roofs [2].

The utilization of three-dimensional information is crucial in various fields, including the medical domain [3]. Hassner et al. resented a novel solution to the problem of depth reconstruction from a single image. 3D reconstruction from a single view is inherently challenging. The authors addressed this issue with a synthesis approach based on examples, involving the use of a database containing mappings from visual appearance to depth for objects within a specific class (e.g., hands, human figures). Their approach entails combining an image of a novel object with the actual depths of patches from similar objects to generate a plausible depth estimate [4].

Patterned textures are effective for conveying 3D images due to two key ingredients: texel distortion and texel distortion rate variation within the texture region. This is also known as a texture gradient. The output of shape-from-texture algorithms is generally a dense map of surface normal. With a smooth textured surface, this is feasible for recovering 3D shapes [5, 6]. Machine learning, machine vision, and image processing have been used to identify manufacturing defects and avoid obstructions in robotics. RGB and depth cameras are not only used for taking pictures but also as nondestructive sensors and robotic path planning [7-9]. 2D to 3D conversion algorithms based on image processing can be split into two groups according to the number of input images: algorithms that use more than one image and algorithms that use one image. For algorithms based on two or more images, typically, more than one input image is taken from a single camera for moving objects in the scene or by multiple fixed cameras at different viewing angles. First-group depth cues are called binocular or multi-ocular depth cues. In the second group of depth cues, monocular depth cues are based on a single still image [6].

Binocular disparity, motion, defocus, focus, and silhouette serve as depth cues in binocular or multi-ocular systems. Binocular disparity, for instance, is employed to ascertain an object's depth by comparing two images captured from slightly different viewpoints of the same scene. Objects closer to the viewing camera exhibit more rapid movement across the retina compared to those farther away, providing a crucial cue for depth perception. Depth-from-defocus methods generate a depth map based on the degree of blurring present in an image.

In contrast, depth-from-focus requires a series of images of the scene with varying focus levels, attained by adjusting the distance between the camera and the scene. On the other hand, depth-from-defocus relies on two or more images with fixed camera and object positions but different focal settings. Silhouettes in images delineate the contours that separate objects from their backgrounds. To employ shape-from-silhouette methods effectively, it becomes necessary to capture multiple views of the scene from different perspectives [6, 10, 11].

Before diving into the complex topic of monocular vision, one must first explore the fundamental depth cues that shape the perception of the visual environment. The following aspects are the depth cues of the monocular: defocus, linear perspective, atmosphere scattering, shading, patterned texture, symmetric patterns, occlusion, statistical patterns. The algorithms in this group are based on depth-from-defocus from a single image, which requires images taken utilizing different focal settings. According to linear perspective, parallel lines, such as rail tracks, appear to merge with distance, eventually vanishing at the horizon. In general, the more converged the lines, the further they seem to be [8]. Through a diffusion of radiation in the atmosphere, light propagation through the atmosphere is affected in the sense that its direction and power are altered. This effect creates the phenomenon known as atmosphere scattering or haze, resulting in various visual effects, such as distant objects appearing bluish and flashlight beams becoming diffused. Object shapes are encoded in the image by the gradual variation of surface shading. Two key factors contribute to a good 3D impression with patterned textures: texel distortions and changes in texel distortion across texture regions. Natural or man-made scenes often feature symmetrical patterns. By using symmetric patterns, 3D reconstruction is possible based on two images of a bilaterally symmetric object from different angles. Objects that overlap or partially obscure another object are considered closer according to depth-from-occlusion algorithms. A statistical pattern is a pattern that occurs repeatedly in images. Machine learning techniques can be effective when input data is large in number or dimension [6, 12, 13].

In a study, edge detection, image segmentation, and feature extraction were used to identify the object and determine its dimensions. The object is then divided into infinitesimally thin slices along the horizontal axis, and the volume is calculated for each slice. The volume of the object is estimated by adding all the volumes of these slices. According to the paper, the proposed method requires only 2 to 3 images depending on the type of object, which is an advantage over other methods that require a minimum of five images. In comparison to the study presented, this system is not autonomous [14]. A novel method was presented in a study in which the camera can freely move around the object, and the focus can be changed independently for each camera view. This method creates a 3D model of an object by estimating its shape-from-silhouette and mapping its texture to the original camera views [15].

Navigating through different or intricate environments becomes challenging, especially when cameras are required to move. This study adopts a straightforward approach, utilizing only two fixed cameras from opposing sides. The primary objective of this research is to investigate the reconstruction of general real-world objects from a limited number of images without prior knowledge of camera pose or object category. This serves as an initial step towards object reconstruction in diverse environments. The significance of this work lies in its unified solution to two fundamental problems in 3D vision: shape reconstruction and pose estimation. Through experiments conducted on both real and synthetic datasets, as well as wild images, the study reveals that achieving reliable object reconstruction necessitates a minimum of five views [16]. Comparatively, the proposed method used in that study only required two views for 3D volume calculation.

Traditionally, depth information has predominantly been a focus in constructing 3D models or extracting details from 2D images within the literature. This paper introduces a novel method that employs an R-C-P approach, implemented in Python and OpenCV, for determining both rectangular surface area and volume from two 2D cameras. Utilizing image processing techniques, including edge detection, the R-C-P method demonstrates consistent results regardless of edge continuity. To establish the correlation between the R-C-P output and actual dimensions, data was collected under varied conditions, excluding instances where the camera-to-object distance remained constant. Interestingly, the results reveal increased error, particularly when maintaining a fixed distance between the camera and objects. Consequently, the study proposes a new method for surface or volume monitoring with a simple camera setup.

The subsequent material and method sections elaborate on the approach employed, detailing the working principle of the R-C-P algorithm. The results and discussion sections provide insights into how R-C-P was applied, and data collected from three rectangular objects, aiming to uncover the relationship between actual and output dimensions. In conclusion, the research outcomes, limitations, and future directions are presented, offering a comprehensive understanding of the study's contributions and potential areas for further exploration.

## 2. Materials and Methods

The primary data used in image processing applications are 2D images or RGB images. An RGB image is a two-dimensional array of data where pixels can be arranged by coordinates as the images intensify [17]. After capturing RGB images, the algorithm converts those pixels into gray-scale images. In addition, the R-C-P algorithm is developed using binary images once edges are detected.

In Figure 1, a flow chart showcases the proposed method before and after applying the R-C-P algorithm. In the beginning, the two cameras capture the RGB image data from the two surfaces of a rectangular object. Next, RGB data was converted into GRAY scale and edge detection before applying the R-C-P approach. Once the R-C-P approach is applied, the detected surface area along with the length and width of the two focal points are then utilized to calculate the volume of the object.

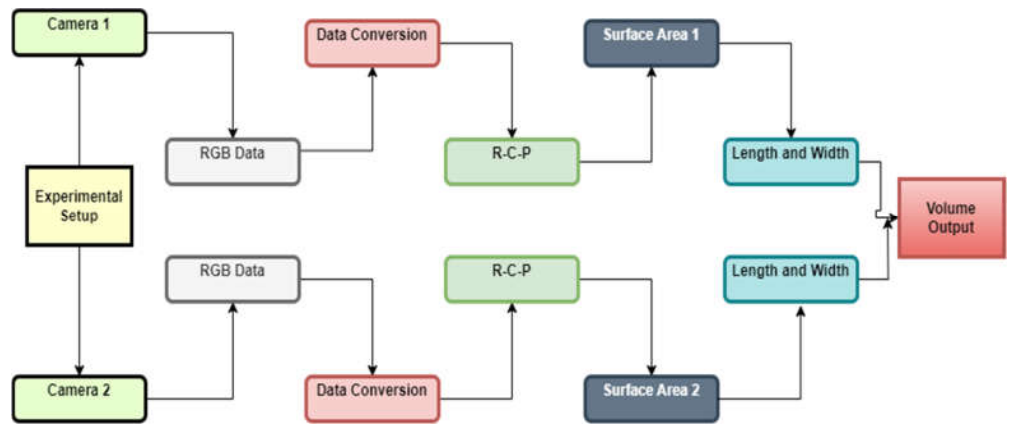

**Figure 1.** Method flow chart.

*2.1. Experimental setup*

To obtain the pictures of the two surfaces of a rectangular object, two cameras have been considered. Figure 2 illustrates an experimental setup using two Intel RealSense cameras.

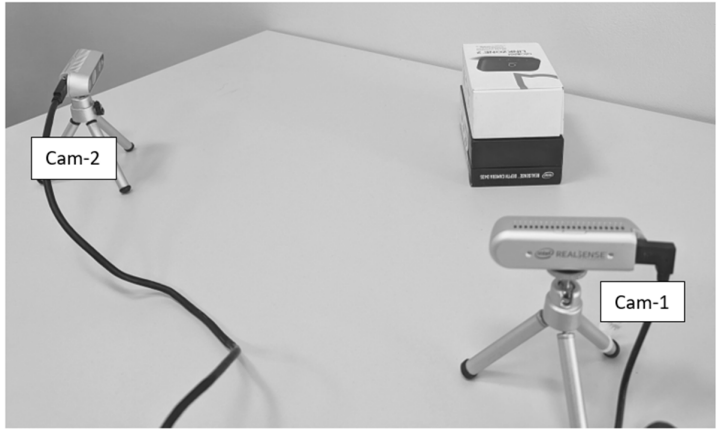

**Figure 2.** Experimental setup.

*2.2. RGB Data and data conversion.*

RGB data capture is the first step after setting up the two cameras from two sides of a rectangular object. After that process, both pictures from Cam-1 and Cam-2 were converted to GRAY scale images. These GRAY scale images are further processed to find the edges.

2.2.1. RGB to GRAY scale

RGB image(s) consists of three colors: red, green, and blue (Figure 3). Each color contains 24 bits per pixel - 8 bits per color band (red, green, blue). The smallest unit of each color is the pixel, which has 256 values (0-255). The color of an image can be modified by varying the number of units in each color band. For example, if we consider a pixel of each channel in a specific location (m, n), where '*m*' indicates the vertical direction location and '*n*' indicates the horizontal directions and set a value of 0 for the red channel, 255 for the green channel, and 0 for the blue channel. If we overlap these channels, then the output image will be a solid green image as the intensity of the green channel is the maximum. For a value of 255 for the red channel, 0 for the green channel, and 0 for the blue channel, the output will be red. The 0 to 255 of all these channels make the unique color of an RGB image. GRAY scale images, on the other hand, are represented by intensity values. There are many shades of GRAY between black and white in GRAY scale images. Each pixel's intensity is based on a range between zero and one (minimum and maximum) and on varying shades of GRAY that range between 0 and 255 [17].

As part of the image processing, the RGB images captured by the camera were first converted to GRAY scale images, as shown in Figure 4.

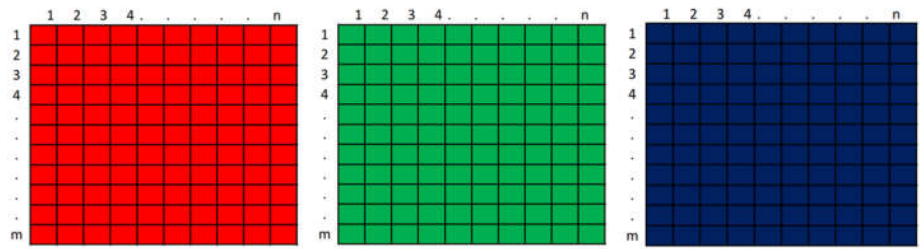

**Figure 3.** Left to right, RGB pixels are represented as Red, Green, and Blue.

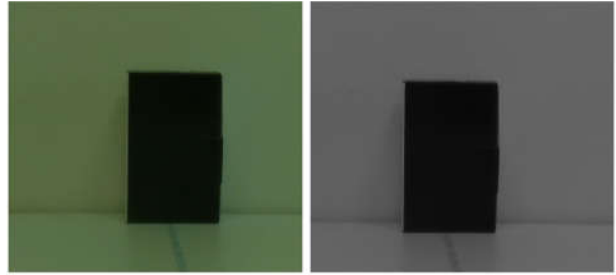

**Figure 4.** The RGB image (left) and the GRAY scale image (right).

2.2.2. Edge detection

The edges of images refer to pixels whose GRAY levels change suddenly, which is the most basic feature of an image. Based on GRAY discontinuous points, edge detection is a basic method for recognizing and segmenting images. Various methods can be utilized for image edge detection, including the Sobel Operator, Prewitt Operator, Canny Edge Detector, Laplacian of Gaussian (LoG), Robert Operator, Marr-Hildreth Edge Detector, and Gradient-Based Methods. Although Canny's edge detection algorithm is more computationally demanding than Sobel, Prewitt, and Robert's operators, it surpasses them in performance despite its higher cost. In this research, Canny Edge detection is applied [18].

John F. Canny proposed the Canny operator in 1986 as a multiply-scale algorithm for edge detection. It is widely used in the field of image processing [19].

The steps for the Canny edge detection are usually divided into several sections:
- Usually, the Gauss smoothing filter is used to reduce noise in the image before detecting the edge.
- Calculate the gradient magnitude and orientation using finite-difference approximations for the partial derivatives. The gradient direction usually corresponds to the four angles - 0, 45, 90, and 135 degrees.
- Non-maximum suppression, thus removing non-edge pixels, leaving only fine lines.
- Set the hysteresis threshold; a hysteresis threshold requires two thresholds that are either retained or excluded to select the edge.
- When the magnitude of a pixel position exceeds a high threshold, it is reserved as an edge pixel. An excluded pixel is one whose magnitude is below the high threshold. Pixel positions between the two thresholds are reserved only when connected to pixels higher than the high threshold [20-22].

*2.3. R-C-P*

Edges are detected as binary images, which are the simplest types of images and have two values, '0' or '1', for each unit. Since each pixel is represented by one binary digit, a binary image is called a 1-bit/pixel image. This R-C-P method uses this black-and-white unit pixel. Figure 5 shows an example of a binary image after edge detection, where white pixels (unit value is 1) indicate the edges of the images and have 14 units of both row and column directions. For the first step, the R-C-P method considers each white pixel in row directions and converts it into white pixels between two parallel positions. For example, for a row-column position (5,6), it searches any white pixel in its parallel directions. In that location, there is no white pixel parallel to the position (5,6). So, there is no conversion from black to white, as shown in Figure 6. In contrast, the positions between (9,6) and (9,12) are converted into white pixels (Figure 6).

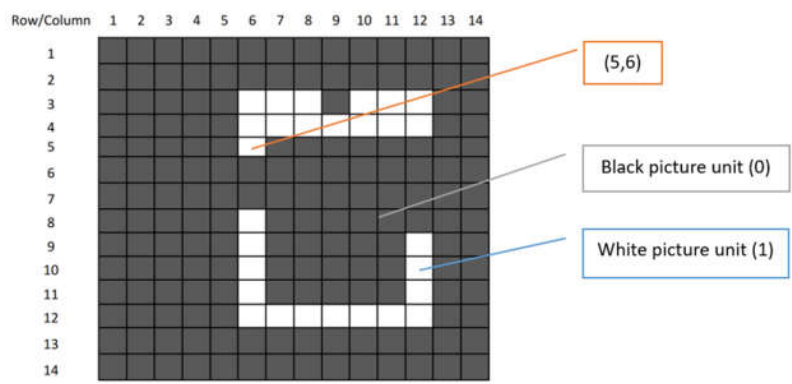

**Figure 5.** Example of an image where white indicates the edges.

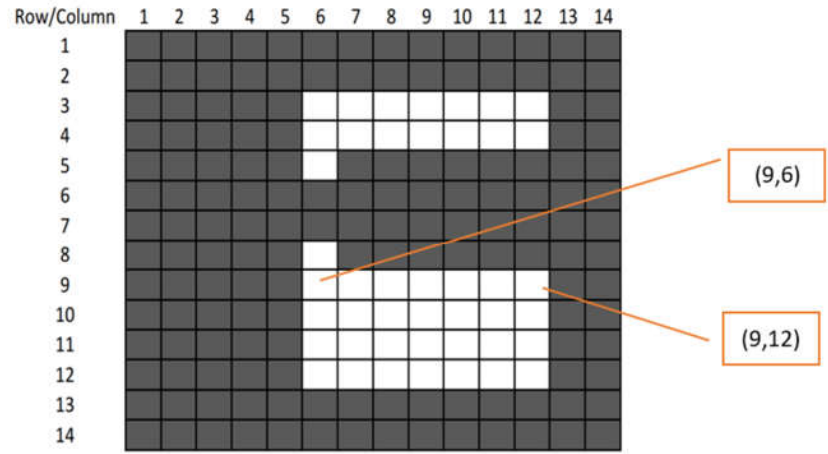

**Figure 6.** R-C-P step 1.

In the second step, the R-C-P method considers each white pixel in column directions and searches for any other white pixel in its parallel (column) direction, and if it detects, then the algorithm converts the black pixel (0) into a white pixel between these two parallel pixels. Figure 7 shows how the black in the position of (6,6) and (7,6) is converted to white by the location of (5,6) and (8,6) being parallel in the vertical direction.

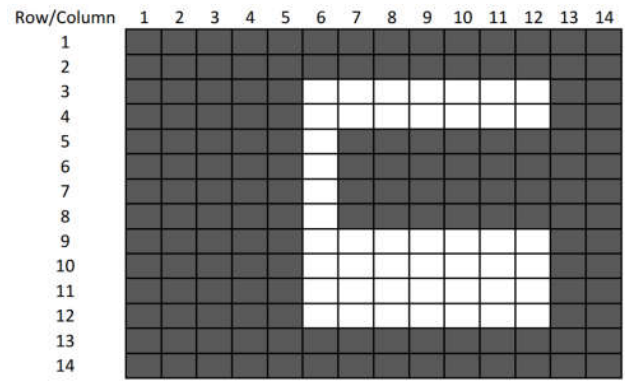

**Figure 7.** R-C-P method step 2.

In Figure 8, the output illustrates the conversion from 0 to 1 is completed after the search in column direction is completed.

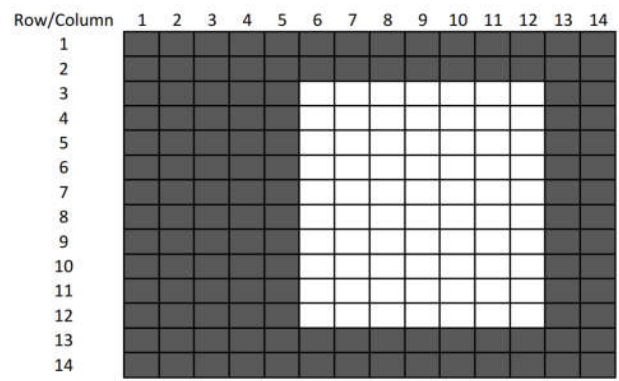

**Figure 8.** R-C-P method output.

*2.4. Surface area to volume output*

In this study, the goal is to obtain the volume of a rectangular object from a camera in real-time. To calculate the area of a surface in the R-C-P method, the total number of white pixels in the output (1 pixel) corresponds to '$d$', where '$d$' is equal to 0.264 mm [23]. Without obtaining at least one dimension (length, width, or height) of the output white pixel area, the surface alone won't yield the volume, even with the deployment of four cameras in all directions. Thus, the width and height calculations are added in the R-C-P method in the final steps.

Two cameras are considered in this study. Assume camera 1 has '$l$' number of white pixels, '$p$' is the number of rows left in the output (1) areas, and '$q$' is the number of columns left in the output (1) areas. In camera 2, '$w$' notates white pixels, '$x$' is the number of rows in the output (1) area, and '$y$' is the number of columns in the output (1) areas.

The detected surface areas have the following width, height, and volume, as shown in equation 1.

For camera 1,
Width/row,   $W1 = (l/p) * d$ mm
Height/column, $H1 = (l/q) * d$ mm

For camera 2,
Width/row,   $W2 = (w/x) * d$ mm
Height/column,   $H2 = (w/y) * d$ mm

The volume of the rectangular object,
$$V = W1 * \{(H1+H2)/2\} * W2 \quad mm^3 \tag{1}$$

In equation 1, the volume is given in real time. As a 2D camera is used as a sensor, this volume may not be the same as the actual dimensions of the object. The experiment presented in the following section uses the R-C-P method to get the relationship between the actual and predicted surfaces.

## 3. Results and Discussion

The method described in the previous section is used in this section to find the result. A variety of distances between the camera and the object were used to collect horizontal and vertical dimensional data (Tables 1 and 2). Data was analyzed to find relationship equations (equations 2 to 7). The comparative results are shown in Table 3.

To check the R-C-P method, an experiment was conducted using three sizes of rectangular objects, as shown in Figure 9. The figure shows the setup that includes a camera and a scale that is used to measure the distance from the surface to the camera face. The objective of this experiment was to calculate and analyze the surface area of the object by varying the distance from the object's surface and the camera distance using the R-C-P method.

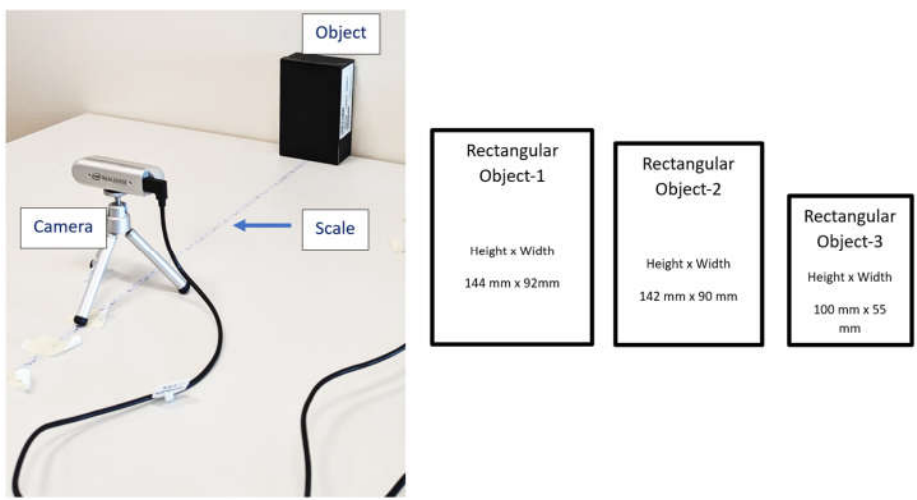

**Figure 9.** Setup and dimensions of the experimental objects.

Distance '$x$' (object surface to camera) varies between 200 mm and 700 mm. R-C-P, however, began collecting results when '$x$' was at 395 mm. This might be because the different cameras have different focus ranges. Figure 10 shows the output after running the Python script. In Figure 10, the picture on the left is the RGB image. The middle of that picture shows the edge detection of the rectangular surface after converting the RGB to gray and applying the edge detection method. The right picture showcases the output of the surface area as a white area by using the R-C-P.

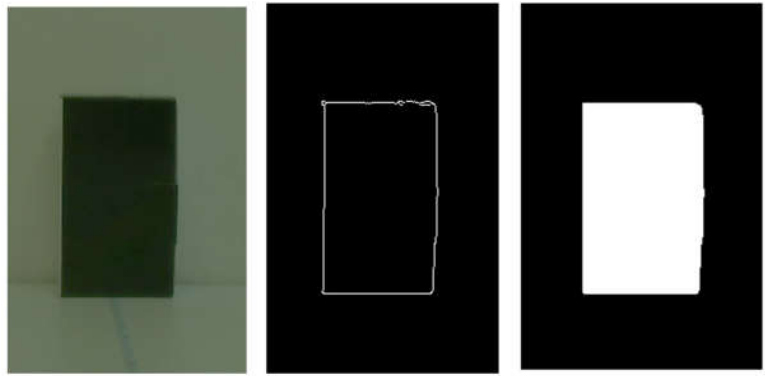

**Figure 10.** The RGB image, edge, and R-C-P methods detected surfaces (left to right).

Canny edge detection faces challenges in accurately detecting edges due to various factors, including the presence of noise, blurry images, complex background patterns, and images with low contrast. Figure 11 illustrates another example of the output by using the R-C-P algorithm. The main difference from the previous example (Figure 10) is the incomplete edge detection (middle picture of Figure 11). However, the output (right image) shows the full surface area of the rectangular object. The reason for this is that R-C-P first checks all row-directional white pixels and then all column-directional white pixels. Therefore, any missing edge does not affect the total surface area. As an example, in Figure 6, location (6,6) has a missing pixel at the edge, but when considered at location (8,6), it is converted to a white pixel in vertical direction.

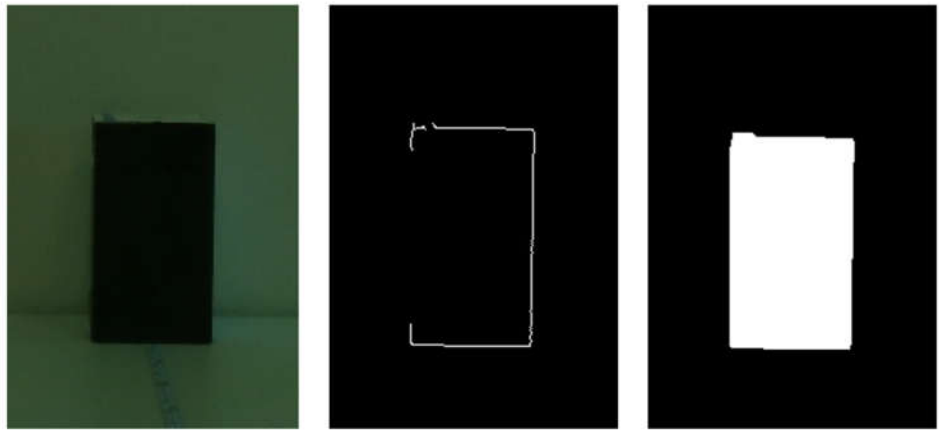

**Figure 11.** R-C-P method output (right), incomplete edge detection (middle), and RGB image (left).

All the objects are placed in the same place by keeping the same distance for the camera surface and applying the method discussed in the previous section from RGB to find the length and width of the surface after getting the output from the R-C-P. Table 1 shows the width (row or horizontal direction of the picture) of objects 1, 2, and 3. Column A of that table shows the distance '$x$'. B, E, and H are the actual width of the objects. Columns C, F, and I show the width calculated from the algorithm of objects 1,2 and 3, respectively. Finally, columns D, G, and J show the ratio of the actual and experimental results for objects 1, 2, and 3, respectively, which indicates the actual dimension of the object is higher than the calculated dimension.

**Table 1.** Actual and R-C-P widths of the target image (objects 1, 2, and 3) surfaces with respect to distances '$x$'.

| | Object -1 | | | Object -2 | | | Object -3 | | |
|---|---|---|---|---|---|---|---|---|---|
| A | B | C | D | E | F | G | H | I | J |
| $x$ (mm) | width/row (mm) | R-C-P method (mm) | D= B/C | width/row (mm) | R-C-P method (mm) | G=E/F | width/row (mm) | R-C-P method (mm) | J=H/I |
| 395 | 55.00 | 6.87 | 8.00 | 90.00 | 15.85 | 5.68 | 92.00 | 17.16 | 5.36 |
| 445 | 55.00 | 5.41 | 10.17 | 90.00 | 12.60 | 7.14 | 92.00 | 13.60 | 6.76 |
| 495 | 55.00 | 4.28 | 12.86 | 90.00 | 10.42 | 8.64 | 92.00 | 10.89 | 8.44 |
| 545 | 55.00 | 3.62 | 15.20 | 90.00 | 8.51 | 10.58 | 92.00 | 9.01 | 10.22 |
| 595 | 55.00 | 3.01 | 18.26 | 90.00 | 7.25 | 12.41 | 92.00 | 7.70 | 11.95 |
| 645 | 55.00 | 2.55 | 21.58 | 90.00 | 5.93 | 15.18 | 92.00 | 6.43 | 14.30 |

The following Table 2 shows the height (column or horizontal direction of the picture) of objects 1, 2, and 3. Column K shows the distance '$x$'. L, O, and R are the actual height of the object. Columns M, P, and S show the width calculated from the algorithm of the objects 1, 2, and 3, respectively. Finally, columns N, Q, and T show the ratio of the actual and experimental results for objects 1, 2, and 3, respectively, which indicates the actual dimension of the object height is higher than the calculated dimension.

**Table 2.** Actual and R-C-P height of the target image (objects 1, 2, and 3) surfaces with respect to distances '$x$'.

| | Object-1 | | | Object -2 | | | Object -3 | | |
|---|---|---|---|---|---|---|---|---|---|
| K | L | M | N | O | P | Q | R | S | T |
| $x$ (mm) | height/column (mm) | R-C-P method (mm) | N=L/M | height/column (mm) | R-C-P method (mm) | Q=O/P | height/column (mm) | R-C-P method (mm) | T=R/S |
| 395 | 100 | 5.15 | 19.40 | 142.00 | 11.89 | 11.95 | 144.00 | 12.87 | 11.19 |
| 445 | 100 | 4.06 | 24.65 | 142.00 | 9.45 | 15.03 | 144.00 | 10.20 | 14.12 |
| 495 | 100 | 3.21 | 31.18 | 142.00 | 7.81 | 18.17 | 144.00 | 8.17 | 17.62 |
| 545 | 100 | 2.71 | 36.84 | 142.00 | 6.38 | 22.26 | 144.00 | 6.75 | 21.32 |
| 595 | 100 | 2.26 | 44.28 | 142.00 | 5.44 | 26.10 | 144.00 | 5.78 | 24.94 |
| 645 | 100 | 1.91 | 52.31 | 142.00 | 4.45 | 31.94 | 144.00 | 4.82 | 29.85 |

Both Tables 1 and 2 indicate that the distance between the object and the camera. This is related to the result of the ratio of the actual and calculated dimensions. The following Figures 12 and 13 represent the data from Tables 1 and 2. Data is taken as the independent value and ratio of the actual dimensions. The calculated dimensions are taken as the dependent value. In both cases, the line of y (D, G, J) and y (N, Q, T) increases as distance '$x$' increases.

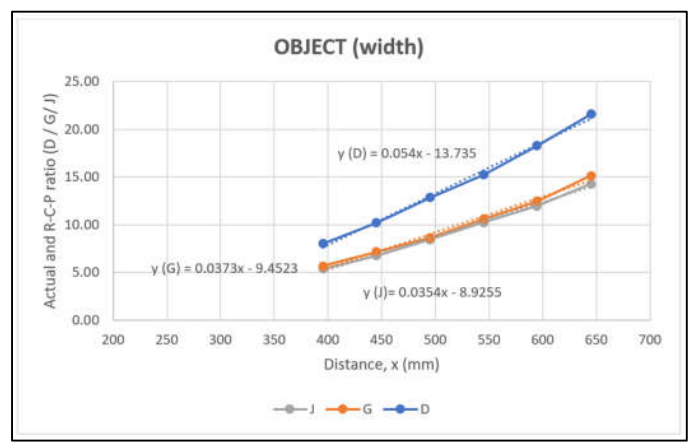

**Figure 12.** Distance versus ratio (D/G/J).

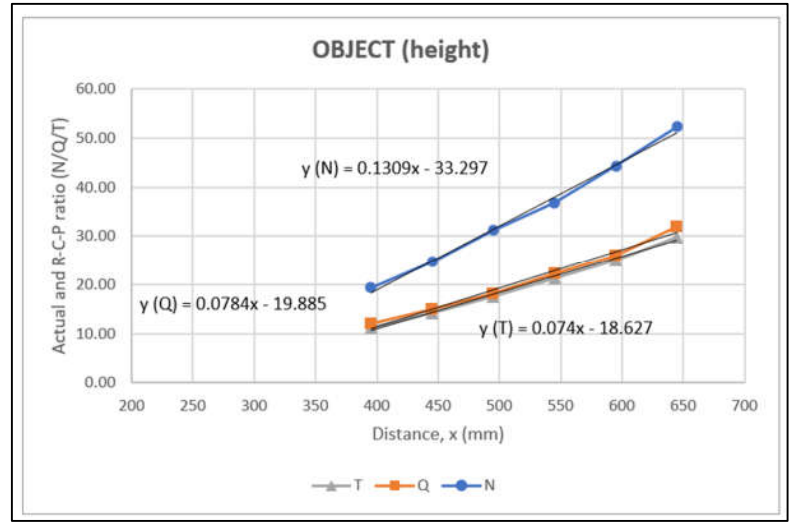

**Figure 13.** Distance versus ratio (T/Q/N).

To find the relationship between the actual and calculated dimension after applying the R-C-P needs further analysis. Based on the data from Table 1 and Figure 12, equations 2, 3 and 4 can be derived. Below, equation 4 shows that r1 represents the ratio of width to row direction where '$x$' represents distance and '$b$' represents the width of the detected surface from the R-C-P method. From the given distance '$x$', from these equations 2, 3, and 4, r1 can be calculated, which makes it possible to find the actual width dimension of the object.

The equation for getting slope:
$$m1 = -0.0064x + 2.4076 \quad (2)$$
The equation for getting intercept:
$$b1 = 0.0661x - 16.807 \quad (3)$$
The ratio of the actual to R-C-P method in the width direction.
$$r1 = m1 * b + b1 \quad (4)$$

Based on the data from Table 2 and Figure 13, equations 5, 6, and 7 can be derived. Equation 7 shows that r2 represents the ratio of width to row direction, and '$x$' represents distance. In the equation, '$h$' represents the height of the detected surface from the R-C-P method. From the given distance '$x$', along with equations 5, 6, and 7, r2 can be calculated, increasing the accuracy of calculating the actual height dimension of the object.

The equation for getting slope:
$$m2 = -0.0265x + 9.9751 \quad (5)$$
The equation for getting intercept:
$$b2 = 0.1681x - 42.735 \quad (6)$$
The ratio of the Actual to R-C-P method in the height direction.
$$r2 = m2 * h + b2 \quad (7)$$

To get the error from the dimension calculation of the rectangular object from equation 4 and 7, a comparison of the ratio object dimensions calculated from the R-C-P output is illustrated in Table 3. Left three columns are catered for the width of the object, while the right three columns are catered for the object's height.

**Table 3.** A comparison of the ratios of the object dimensions calculated from R-C-P and equations.

| Surface width/row | | | Surface height/column | | |
|---|---|---|---|---|---|
| Objects width/width from the output of R-C-P | Objects width/width from equation 4 | Error (%) | Objects height/height from the output of R-C-P | Objects height/height from equation 7 | Error (%) |
| 8.48 | 8.00 | 5.90 | 21.13 | 19.40 | 8.89 |
| 7.39 | 5.68 | 30.21 | 17.81 | 11.95 | 49.09 |
| 7.24 | 5.36 | 34.95 | 17.33 | 11.19 | 54.83 |
| 10.23 | 10.17 | 0.55 | 24.70 | 24.65 | 0.18 |
| 7.06 | 7.14 | 1.19 | 14.90 | 15.03 | 0.87 |
| 6.62 | 6.76 | 2.17 | 13.53 | 14.12 | 4.15 |
| 12.66 | 12.86 | 1.55 | 30.39 | 31.18 | 2.51 |
| 7.99 | 8.64 | 7.50 | 15.92 | 18.17 | 12.40 |
| 7.63 | 8.44 | 9.67 | 14.80 | 17.62 | 16.03 |
| 15.31 | 15.20 | -0.73 | 36.75 | 36.84 | 0.23 |
| 10.03 | 10.58 | 5.23 | 20.38 | 22.26 | 8.44 |
| 9.49 | 10.22 | 7.13 | 18.70 | 21.32 | 12.27 |
| 18.31 | 18.26 | 0.22 | 44.20 | 44.28 | 0.17 |
| 12.36 | 12.41 | 0.35 | 25.77 | 26.10 | 1.26 |
| 11.74 | 11.95 | 1.75 | 23.83 | 24.94 | 4.42 |
| 21.44 | 21.58 | 0.63 | 52.08 | 52.31 | 0.44 |
| 15.63 | 15.18 | 2.94 | 34.05 | 31.94 | 6.60 |
| 14.76 | 14.30 | 3.20 | 31.36 | 29.85 | 5.04 |

According to the data, excluding the distance of 395 mm, the experimental study reveals errors of up to 10 percent for the width or row, typically staying below 4 percent. Conversely, excluding the distance 'x' of 395 mm results in errors for the height or column length of the detected surface, reaching up to 17 percent, with the majority of errors below 7 percent.

The primary objective of this data analysis is to assess the accuracy of length and width calculations derived from the equations based on the data presented in Tables 1 and 2. The subsequent subsection details the process of obtaining volume calculations using the R-C-P approach and the equations developed in this study.

*3.1. Calculation of the volume from the camera*

Based on two cameras, let us assume the width of the surface using the R-C-P method is '$x1$' and '$x2$' for camera 1 and camera 2. The height or column for cameras 1 and 2 from the R-C-P method is '$y1$' and '$y2$', respectively.

We can get *m1, m2, b1,* and *b2* from equations 2, 3, 5, and 6 if we know the distance of *'x'*, from equations 4 and 7.

Camera 1:  $r11 = m1 * x1 + b1$
$r21 = m2 * y1 + b2$
Camera 2:  $r12 = m1 * x2 + b1$
$r22 = m2 * y2 + b2$

Object width:
Camera 1:  $w1 = r11 * x1$
Camera 2:  $w2 = r12 * x2$

Object height:
Camera 1:  $h1 = r21 * y1$
Camera 2:  $h2 = r22 * y2$

Volume of the object is $w1 * (h1+h2)/2 * w2$.

## 4. Conclusion

This study introduces a volume calculation method using image processing named R-C-P to estimate the volume of a rectangular object without a LiDAR or 3D depth camera. Subsequently, data collected through R-C-P is analyzed to derive equations for calculating the detected surface width and column length, contingent upon the distance between the object and the camera.

To the best of our knowledge, as indicated by the literature reviews, the R-C-P method presented here represents a unique approach to volume calculation. The primary findings of this study underscore that, although the image processing method may not precisely identify the finalized edge of a surface, it remains effective in determining the overall surface area. Despite inherent errors in calculating length, width, and volume using the derived equations, this method demonstrates its capability in monitoring volume changes within a system that necessitates such information, all without relying on a depth sensor.

All research studies have limitations, but this study is no different. The results reveal a higher error in the equations up to a specific distance, potentially attributed to the lack of proper camera calibration and varying environmental conditions, including lighting. Both factors exert a significant influence on image processing. Future studies will focus on a thorough calibration process and compare it with the current setup. Additionally, environmental effects, such as adjusting light intensity and location, will be systematically analyzed to minimize errors in calculations.

To further enhance accuracy and address potential sources of error, future research will incorporate variations in object shapes, including square and circular geometries, considering that only inaccurate rectangular shapes were used in the current study. Exploring the impact of the camera itself as a potential source of error, higher resolution cameras will be employed in future investigations to assess their influence and compare results with the present study. These considerations collectively contribute to a comprehensive understanding and improvement of the proposed methodology.

**Acknowledgment:** The authors would like to express their sincere gratitude to the anonymous reviewers for their valuable comments and constructive feedback, which significantly improved the quality of this article.